\documentclass[10pt,twocolumn,twoside]{article}
\usepackage[
  letterpaper,
  top=2.5cm,
  bottom=2.5cm,
  left=19mm,
  right=19mm,
  footskip=1.7cm
]{geometry}

\usepackage{titling}

\usepackage{newtxtext,newtxmath}
\usepackage{graphicx,booktabs,array,amsmath,microtype,cite,fancyhdr,url}
\usepackage[font=small,labelfont=bf,labelsep=colon]{caption}
\usepackage{dblfloatfix}
\usepackage[normalem]{ulem}
\graphicspath{{figures/}}
\makeatletter
\renewcommand\section{\@startsection{section}{1}{\z@}{2.6ex plus .6ex minus .2ex}{1.6ex plus .4ex}{\centering\normalfont\normalsize\bfseries\MakeUppercase}}
\renewcommand\subsection{\@startsection{subsection}{2}{\z@}{1.0ex plus .2ex minus .2ex}{.4ex}{\normalfont\normalsize}}
\makeatother
\usepackage{eso-pic}

\AddToShipoutPictureFG{%
  \ifodd\value{page}%

    \AtPageUpperLeft{%
      \put(\LenToUnit{0.5\paperwidth},
           \LenToUnit{-4mm}){%
        \makebox(0,0)[ct]{%
          \includegraphics[
            height=18mm,
            keepaspectratio
          ]{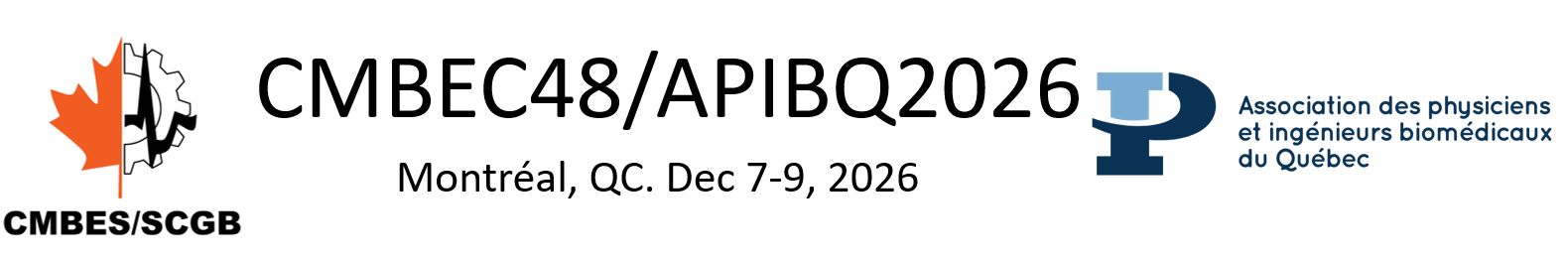}%
        }%
      }%
    }%

    \AtPageUpperLeft{%
      \put(
        \LenToUnit{\dimexpr\paperwidth-19mm\relax},
        \LenToUnit{-24mm}
      ){%
        \makebox(0,0)[rt]{\thepage}%
      }%
    }%

  \else

    \AtPageUpperLeft{%
      \put(
        \LenToUnit{19mm},
        \LenToUnit{-17mm}
      ){%
        \makebox(0,0)[lt]{\thepage}%
      }%
    }%
  \fi
}

\title{\bfseries\fontsize{14}{16}\selectfont LAF-ECG: Multi-Resolution Lead-Aware Fusion of Time-Series and Language Foundation Models for ECG Classification}
\author{\fontsize{12}{14}\selectfont Kiana Taktehrani$^{1}$, Maryam Hosseini$^{1}$, Amir Keynia$^{2}$, Aram Montazami$^{2}$, and Arash Mohammadi$^{1}$\\[3pt]
{\itshape\fontsize{9}{11}\selectfont $^{1}$Cybersecurity and Intelligent Systems Engineering, Concordia University, Montreal, Canada}
\\ {\itshape\fontsize{9}{11}\selectfont $^{2}$Novatek International Network of Corporation, Montreal, Canada}}
\date{}

\begin{document}

\maketitle
\thispagestyle{fancy}
\vspace{-1.5em}

\begin{abstract}
\noindent Automated electrocardiogram (ECG) interpretation has advanced substantially with deep learning, yet many methods remain task-specific and label-dependent. Large language models (LLMs) and time-series foundation models (TSFMs) offer complementary pretrained representations, but integrating them is challenging because of mismatched feature spaces and the lead- and scale-dependent structure of ECG signals. We propose LAF-ECG (Lead-Aware Foundation Fusion for ECG), a dual-foundation framework that extends MedTsLLM with a lead-aware interface to a frozen TSFM backbone while retaining its reprogramming pathway. The proposed framework captures multi-resolution temporal context, fuses temporal and language-aligned ECG representations through adaptive residual integration, and employs branch-specific supervision with cross-branch alignment. LAF-ECG achieves 71.10\% accuracy and 62.43\% macro-F1 on PTB-XL, and 89.73\% accuracy and 65.94\% macro-F1 on Chapman--Shaoxing, outperforming MedTsLLM on both datasets. These results highlight the value of preserving lead-specific temporal structure when combining temporal and language foundation representations for ECG classification. The code is publicly available at \url{https://github.com/kkianamm/LAF-ECG}.
\end{abstract}


\section*{INTRODUCTION}

The electrocardiogram (ECG) is a widely used non-invasive tool for cardiac assessment, with diagnostic information distributed across multiple leads and temporal patterns. Although deep learning has achieved strong ECG classification performance \cite{strodthoff2020deep}, many methods remain task-specific and annotation-dependent, limiting the transferability of learned representations.

Foundation models offer a promising alternative. Large-scale ECG pretraining, as demonstrated by ECG-FM \cite{mckeen2025ecgfm}, can provide reusable physiological representations, while LLM-based reprogramming maps numerical time series into pretrained language-model spaces. Time-LLM introduced this paradigm for general time series \cite{jin2024timellm}, and MedTsLLM extended it to physiological signals \cite{chan2024medtsllm}. In parallel, time-series foundation models (TSFMs), including MOMENT \cite{goswami2024moment}, provide strong pretrained temporal representations, and Time2Lang has explored combining TSFMs with LLMs for health sensing \cite{pillai2025time2lang}. However, effectively integrating these complementary representations while preserving lead-specific and multi-resolution ECG information remains challenging.

We propose \textbf{LAF-ECG} (Lead-Aware Foundation Fusion for ECG), a dual-foundation framework that combines the MedTsLLM reprogramming pathway with a frozen MOMENT-1-base temporal backbone \cite{goswami2024moment}. A Lead-Aware Temporal Foundation Module (LTFM) models token-dependent inter-lead importance, while Multi-Resolution Temporal Context (MRTC) captures complementary global information from overlapping ECG windows. Adaptive Residual Token Fusion (ARTF) selectively integrates temporal and reprogrammed representations, and Branch-Specific Auxiliary Supervision (BAS) together with Supervised Cross-Branch Alignment (SCA) encourages discriminative and class-consistent features across both pathways.

We evaluate LAF-ECG on PTB-XL \cite{wagner2020ptbxl} and Chapman--Shaoxing \cite{zheng2020chapman}. Experiments demonstrate consistent improvements over the evaluated baselines, while ablation studies confirm the complementary contributions of the proposed components. Our main contributions are: (1) a lead-aware, multi-resolution interface for adapting a pretrained TSFM to ECGs; (2) adaptive residual fusion of temporal and MedTsLLM representations; and (3) joint branch supervision and alignment for effective ECG classification with frozen temporal and language foundation backbones.

\begin{figure*}[t]
\centering
\includegraphics[width=0.78\textwidth]{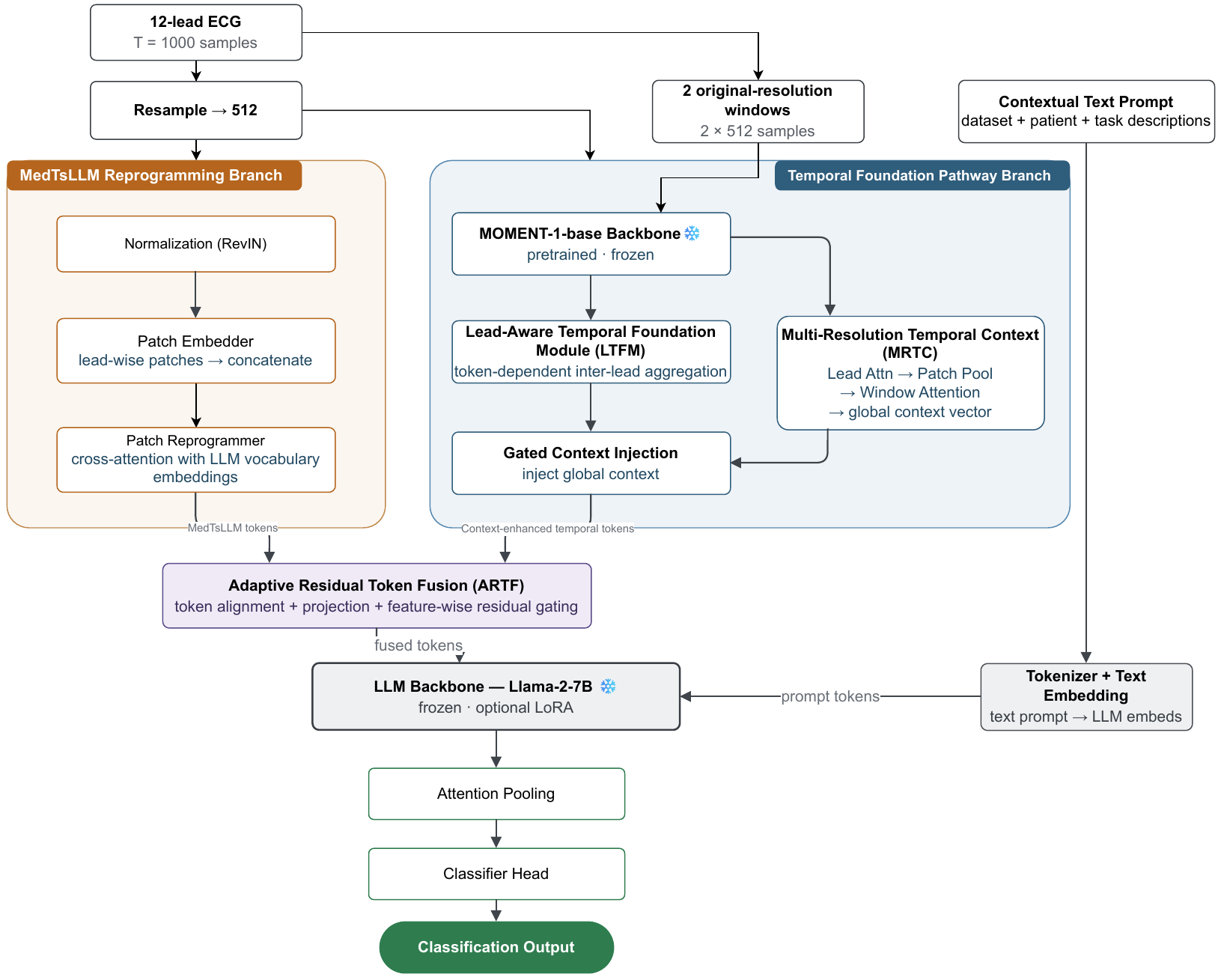}
\caption{\sloppy Overview of the proposed LAF-ECG architecture, including the MedTsLLM reprogramming branch, lead-aware temporal foundation pathway, multi-resolution temporal context, and adaptive residual token fusion.}
\label{fig:architecture}
\end{figure*}

\section*{METHODOLOGY}

LAF-ECG consists of two complementary pathways: the MedTsLLM reprogramming branch \cite{chan2024medtsllm} and a lead-aware temporal branch built on frozen MOMENT \cite{goswami2024moment} as shown in Figure 1. Given a standardized 12-lead ECG
$X^{\mathrm{orig}}\in\mathbb{R}^{1000\times C}$, where $C=12$, the input is temporally resampled from 1000 to 512 samples for the aligned pathway:
\begin{equation}
X^{\mathrm{align}}
=\operatorname{Resample}(X^{\mathrm{orig}},512)
\in\mathbb{R}^{512\times C}.
\end{equation}
The original-resolution signal is retained for multi-resolution temporal context. The model predicts $y\in\{1,\ldots,K\}$, while MOMENT-1-base and Llama-2-7B \cite{touvron2023llama2} remain frozen. Let $N_m$ and $N_o$ denote the MedTsLLM and MOMENT token counts, respectively, and $d_p$, $d_o$, and $d_L$ denote the patch, MOMENT, and LLM embedding dimensions. Throughout, uppercase symbols such as $W$ denote learnable matrices, while lowercase symbols such as $w$, $v$, and $b$ denote learnable vectors or biases as specified in context. We use $[\cdot;\cdot]$ for concatenation, $\odot$ for element-wise multiplication, $\sigma(\cdot)$ for the sigmoid function, and $\operatorname{LN}(\cdot)$ for LayerNorm.

\subsection*{\underline{Dual-Path ECG Representation}}

\textit{MedTsLLM reprogramming branch.}
Following MedTsLLM \cite{chan2024medtsllm}, $X^{\mathrm{align}}$ is normalized by RevIN, divided into 8-sample patches with stride 8, and embedded lead-wise to obtain $P\in\mathbb{R}^{N_m\times Cd_p}$. The frozen vocabulary embeddings $E_{\mathrm{vocab}}\in\mathbb{R}^{V\times d_L}$ are mapped by the learnable projection $W_m$ to source prototypes $E_s=W_mE_{\mathrm{vocab}}$. Reprogramming is performed by
\begin{equation}
H^{\mathrm{med}}
=
\operatorname{softmax}\!\left(
\frac{PW_Q(E_sW_K)^\top}{\sqrt{d_k}}
\right)E_sW_VW_O ,
\end{equation}
where $W_Q$, $W_K$, $W_V$, and $W_O$ are the learnable query, key, value, and output projections, respectively, and $d_k$ is the per-head query/key dimension. Thus,
$H^{\mathrm{med}}=[h^{\mathrm{med}}_1,\ldots,h^{\mathrm{med}}_{N_m}]
\in\mathbb{R}^{N_m\times d_L}$,
where $h_n^{\mathrm{med}}$ is the $n^{\text{th}}$ language-aligned ECG token.

\textit{Lead-Aware Temporal Foundation Module (LTFM).}
Frozen MOMENT returns unreduced lead-wise tokens
$H^{\mathrm{mom},0}\in\mathbb{R}^{C\times N_o\times d_o}$.
For each lead $c$, a learnable embedding $e_c\in\mathbb{R}^{d_o}$ is added, and token-dependent lead weights are computed as
\begin{align}
a_{c,n}
&=w_a^\top\tanh\!\left(
W_a(H^{\mathrm{mom},0}_{c,n}+e_c)\right),\\
\alpha_{c,n}
&=\frac{\exp(a_{c,n})}
{\sum_{j=1}^{C}\exp(a_{j,n})},\qquad
z_n=\sum_{c=1}^{C}\alpha_{c,n}
(H^{\mathrm{mom},0}_{c,n}+e_c),
\end{align}
where $W_a$ and $w_a$ parameterize the lead-attention scorer. This yields
$Z=[z_1,\ldots,z_{N_o}]\in\mathbb{R}^{N_o\times d_o}$.

For compactness, learned attention pooling of
$U=[u_1,\ldots,u_M]$ is written as
\begin{equation}
\operatorname{AP}(U)
=
\sum_j \beta_j u_j,
\qquad
\beta_j
=
\frac{
\exp\!\left(
v_{\mathrm{AP}}^\top
\tanh(W_{\mathrm{AP}}u_j)
\right)}
{
\sum_{\ell}
\exp\!\left(
v_{\mathrm{AP}}^\top
\tanh(W_{\mathrm{AP}}u_\ell)
\right)
}.
\end{equation}
where $W_{\mathrm{AP}}$ and $v_{\mathrm{AP}}$ are learnable scoring parameters, independently defined for each pooling module.

\subsection*{\underline{Multi-Resolution Context and Adaptive Fusion}}

\textit{Multi-Resolution Temporal Context (MRTC).}
Two overlapping original-resolution windows,
$X^{(1)}=X^{\mathrm{orig}}_{1:512}$ and
$X^{(2)}=X^{\mathrm{orig}}_{489:1000}$,
are processed by the same frozen MOMENT backbone and shared LTFM. Their lead-aware tokens $Z^{(w)}$ are summarized by patch and window attention,
\begin{equation}
q_w=\operatorname{AP}_{p}(Z^{(w)}),\qquad
g=\operatorname{AP}_{w}([q_1,q_2]),
\end{equation}
where $g\in\mathbb{R}^{d_o}$ is the global temporal context. It is injected into the aligned temporal tokens by
\begin{equation}
\gamma_n=\sigma(W_g[z_n;g]+b_g),\qquad
\widetilde z_n=
\operatorname{LN}(z_n+\gamma_n\odot g),
\end{equation}
where $W_g$ and $b_g$ are the learnable weight matrix and bias vector of the context gate, and $\gamma_n$ is its feature-wise gating vector.

\textit{Adaptive Residual Token Fusion (ARTF).}
The MedTsLLM token count is aligned to $N_o$ by cropping a single extra padded token when present and otherwise using linear interpolation. Let $\bar h_n^{\mathrm{med}}$ denote the $n^{\text{th}}$ MedTsLLM token after this alignment. The temporal representation is projected to the LLM space,
$\widehat z_n=\phi(\widetilde z_n)\in\mathbb{R}^{d_L}$,
where $\phi(\cdot)$ is a two-layer GELU MLP with dropout and LayerNorm. ARTF computes
\begin{align}
r_n&=\sigma\!\left(
\operatorname{MLP}([\bar h_n^{\mathrm{med}};\widehat z_n])
\right),\\
f_n&=\operatorname{LN}\!\left(
\bar h_n^{\mathrm{med}}+r_n\odot\widehat z_n
\right),
\end{align}
where $r_n$ is a feature-wise fusion gate. The MedTsLLM representation therefore remains the residual anchor, producing
$F=[f_1,\ldots,f_{N_o}]$.

\subsection*{\underline{LLM Contextualization and Training}}

Prompt embeddings
$P_{\mathrm{text}}\in\mathbb{R}^{L_p\times d_L}$
encode dataset, patient/record, and task information and are prepended to $F$; statistics and example prompts are disabled. For PTB-XL, patient metadata include age, sex, and heart axis. Frozen Llama-2-7B processes
$S=[P_{\mathrm{text}};F]$, and only ECG-position hidden states $H_{\mathrm{ECG}}$ are retained and pooled,
\begin{equation}
\eta_n=
\frac{\exp(v_c^\top\tanh(W_cH_{\mathrm{ECG},n}))}
{\sum_j\exp(v_c^\top\tanh(W_cH_{\mathrm{ECG},j}))},
\qquad
h_{\mathrm{ECG}}=\sum_n\eta_nH_{\mathrm{ECG},n},
\end{equation}
where $W_c$ and $v_c$ are learnable pooling parameters. A LayerNorm--Dropout--Linear head maps $h_{\mathrm{ECG}}$ to $K$ logits.

\textit{Branch-Specific Auxiliary Supervision (BAS).}
Independent attention pooling and classifiers are applied to the pre-fusion aligned MedTsLLM tokens and projected MOMENT tokens, producing logits $\ell^{\mathrm{med}}$ and $\ell^{\mathrm{mom}}$:
\begin{equation}
\mathcal{L}_{\mathrm{aux}}
=\frac{1}{2}\left[
\operatorname{CE}(\ell^{\mathrm{med}},y)
+\operatorname{CE}(\ell^{\mathrm{mom}},y)
\right].
\end{equation}

\textit{Supervised Cross-Branch Alignment (SCA).}
Let $u_i$ and $v_i$ be the L2-normalized pooled MedTsLLM and MOMENT representations for sample $i$ in a batch of size $B$, and
$\mathcal{P}(i)=\{j:y_j=y_i\}$ the same-class positive set. Defining
$s_{ij}=u_i^\top v_j/\tau$, the symmetric cross-branch loss is
\begin{equation}
\mathcal{L}_{\mathrm{align}}
=-\frac{1}{2B}\sum_{i=1}^{B}
\frac{1}{|\mathcal{P}(i)|}
\sum_{j\in\mathcal{P}(i)}
\left[
\log\frac{e^{s_{ij}}}{\sum_k e^{s_{ik}}}
+
\log\frac{e^{s_{ji}}}{\sum_k e^{s_{ki}}}
\right].
\end{equation}
The complete objective is
\begin{equation}
\mathcal{L}
=
\mathcal{L}_{\mathrm{cls}}
+\lambda_{\mathrm{aux}}\mathcal{L}_{\mathrm{aux}}
+\lambda_{\mathrm{align}}\mathcal{L}_{\mathrm{align}},
\end{equation}
where $\mathcal{L}_{\mathrm{cls}}$ is class-weighted cross-entropy with label smoothing,
$\lambda_{\mathrm{aux}}=0.20$,
$\lambda_{\mathrm{align}}=0.05$, and
$\tau=0.10$.

\nocite{zhou2023onefitsall,nie2022patchtst,langer2025opentslm,lan2026gem}

\section*{EXPERIMENTS AND RESULTS}
\subsection*{\underline{Datasets}}
Here, we evaluate LAF-ECG only for ECG classification. For PTB-XL \cite{wagner2020ptbxl}, 100-Hz 12-lead recordings are used with the standard patient-wise folds: 1--8 for training, 9 for validation, and 10 for testing. Diagnostic SCP codes are mapped to the five superclasses NORM, MI, STTC, CD, and HYP; likelihoods are summed within each superclass, and the superclass with the largest accumulated score is selected to obtain a single recording-level label. For the Chapman--Shaoxing database \cite{zheng2020chapman}, recordings are downsampled to 100 Hz and an eight-class single-rhythm subset is constructed: SB, SR, AFIB, ST, AF, SA, SVT, and AT. Multi-rhythm records are excluded, and a deterministic stratified 70/10/20 train/validation/test split is used. The Chapman--Shaoxing subset contains 43,021 recordings, split into 30,114/4,302/8,605 train/validation/test samples using seed 0; class counts range from 141 (AT) to 15,824 (SB), indicating substantial imbalance. A 512-sample representation is provided to the aligned MedTsLLM and LTFM pathways, while the normalized 1000-sample ECG is retained to generate two overlapping 512-sample high-resolution windows for the MRTC pathway.

\subsection*{\underline{Experimental Setup}}
The pretrained MOMENT-1-base backbone \cite{goswami2024moment} and Llama-2-7B \cite{touvron2023llama2} are frozen, while the MedTsLLM adaptation layers and the proposed LTFM, MRTC, temporal-to-language projection, ARTF, attention-pooling, BAS, SCA, and classifier are optimized. ECGs are standardized using training-set statistics and augmented during training with $\pm5\%$ per-lead amplitude scaling ($\sigma=0.01$), lead dropout ($p=0.05$), and a 2\% temporal mask. Models are trained for 20 epochs with AdamW (learning rate $3\times10^{-4}$, weight decay 0.01), class-weighted cross-entropy with 0.05 label smoothing, gradient clipping at 1.0, and classifier dropout of 0.20. A batch size of 4 with four-step gradient accumulation gives an effective batch size of 16. Model performance is evaluated on the held-out test set using accuracy, macro-F1, precision, and recall. Experiments are implemented in PyTorch/Hugging Face Transformers and run on an NVIDIA H100 80GB GPU.

LAF-ECG contains approximately 6.77B parameters, of which 52.3M (0.77\%) are trainable. Recorded batched test throughput is approximately 53 ECG/s on PTB-XL and 57 ECG/s on Chapman--Shaoxing; peak GPU memory and batch-size-1 inference latency were not recorded. On PTB-XL, three runs with seeds 0, 42, and 2026 yielded a mean $\pm$ standard deviation of $70.74\pm0.33\%$ accuracy and $61.93\pm0.43\%$ macro-F1, with 95\% t-based confidence intervals of $[69.91,71.56]\%$ for accuracy and $[60.85,63.01]\%$ for macro-F1. For reproduced baselines, held-out evaluation partitions were kept consistent where applicable, while model-specific hyperparameters followed their corresponding implementation settings.

\begin{table}[t]
\centering
\caption{Classification performance on PTB-XL and Chapman--Shaoxing.}
\label{tab:main_results}
\small
\renewcommand{\arraystretch}{0.96}
\setlength{\tabcolsep}{3.3pt}

\begin{tabular*}{\linewidth}{@{\extracolsep{\fill}}lcccc}
\toprule
\textbf{Method} & \textbf{Acc.} & \textbf{F1} &
\textbf{Prec.} & \textbf{Rec.} \\
\midrule

\multicolumn{5}{c}{\textit{PTB-XL}} \\
\cmidrule(lr){1-5}
GPT4TS \cite{zhou2023onefitsall}
    & 57.58 & 40.83 & 48.23 & 40.68 \\
PatchTST \cite{nie2022patchtst}
    & 63.47 & 49.19 & 52.12 & 47.78 \\
OpenTSLM \cite{langer2025opentslm}
    & 65.79 & 41.52 & 44.61 & 40.55 \\
MedTsLLM \cite{chan2024medtsllm}
    & 66.85 & 51.07 & 54.20 & 49.85 \\
GEM \cite{lan2026gem}
    & 69.88 & 51.32 & 54.18 & 49.49 \\
\textbf{LAF-ECG}
    & \textbf{71.10} & \textbf{62.43}
    & \textbf{63.01} & \textbf{62.04} \\

\addlinespace[2pt]
\midrule

\multicolumn{5}{c}{\textit{Chapman--Shaoxing}} \\
\cmidrule(lr){1-5}
GPT4TS \cite{zhou2023onefitsall}
    & 75.60 & 33.97 & 38.46 & 35.45 \\
PatchTST \cite{nie2022patchtst}
    & 75.20 & 34.95 & 47.38 & 35.82 \\
OpenTSLM \cite{langer2025opentslm}
    & 83.62 & 48.78 & 54.00 & 48.36 \\
MedTsLLM \cite{chan2024medtsllm}
    & 85.26 & 59.35 & 61.85 & 58.29 \\
GEM \cite{lan2026gem}
    & 84.97 & 51.56 & 53.76 & 52.06 \\
\textbf{LAF-ECG}
    & \textbf{89.73} & \textbf{65.94}
    & \textbf{64.50} & \textbf{68.47} \\
\bottomrule

\end{tabular*}
\end{table}

\subsection*{\underline{Main Results}}
As shown in Table~1, LAF-ECG achieves the best overall performance on both datasets, with 71.10\% accuracy and 62.43\% macro-F1 on PTB-XL, and 89.73\% accuracy and 65.94\% macro-F1 on Chapman--Shaoxing. Table~1 reports the seed-0 PTB-XL result, while the three-seed aggregate is reported above. These results demonstrate the effectiveness of the proposed lead-aware, multi-resolution fusion framework.

\begin{table}[t]
\centering
\caption{Ablation analysis of LAF-ECG on PTB-XL.}
\label{tab:ablation}
\small
\renewcommand{\arraystretch}{0.96}
\setlength{\tabcolsep}{3.3pt}

\begin{tabular*}{\linewidth}{@{\extracolsep{\fill}}lcccc}
\toprule
\textbf{Model / Variant} & \textbf{Acc.} & \textbf{F1} &
\textbf{Prec.} & \textbf{Rec.} \\
\midrule
MedTsLLM
    & 66.85 & 51.07 & 54.20 & 49.85 \\
w/o High-Res Context
    & 67.91 & 58.36 & 59.65 & 58.59 \\
w/o Alignment
    & 69.30 & 59.92 & 61.44 & 59.81 \\
w/ Mean Lead Pooling
    & 68.89 & 59.33 & 59.82 & 59.04 \\
\textbf{LAF-ECG}
    & \textbf{71.10} & \textbf{62.43}
    & \textbf{63.01} & \textbf{62.04} \\
\bottomrule
\end{tabular*}
\end{table}

\subsection*{\underline{Ablation Study}}
As shown in Table~2, all ablations reduce performance compared with the full LAF-ECG model. Removing high-resolution context causes the largest macro-F1 drop ($-4.07$ points), followed by mean lead pooling ($-3.10$) and removal of cross-branch alignment ($-2.51$). These results confirm that each component contributes to the model's overall performance.

\section*{CONCLUSION}
LAF-ECG is a dual-foundation framework that combines lead-aware temporal modeling with the MedTsLLM reprogramming pathway for ECG classification. By integrating multi-resolution context, adaptive fusion, and cross-branch alignment, it effectively combines frozen time-series and language foundation models while preserving lead-specific information. Experiments on PTB-XL and Chapman--Shaoxing demonstrate consistent improvements over the evaluated baselines, with ablation studies confirming the contribution of the proposed components. These results support lightweight ECG-specific adaptation for foundation-model-based ECG classification. Code and configurations will be publicly released upon acceptance.

{\fontsize{8}{9}\selectfont
 \bibliographystyle{IEEEtran}
 \bibliography{references}

@article{strodthoff2020deep,
  title={Deep learning for {ECG} analysis: benchmarks and insights from {PTB-XL}},
  author={Strodthoff, Nils and Wagner, Patrick and Schaeffter, Tobias and Samek, Wojciech},
  journal={IEEE journal of biomedical and health informatics},
  volume={25},
  number={5},
  pages={1519--1528},
  year={2020},
  publisher={IEEE}
}

@article{mckeen2025ecgfm,
  title= {{ECG-FM}: an open electrocardiogram foundation model},
  author={McKeen, Kaden and Masood, Sameer and Toma, Augustin and Rubin, Barry and Wang, Bo},
  journal={JAMIA open},
  volume={8},
  number={5},
  pages={ooaf122},
  year={2025},
  publisher={Oxford University Press}
}

@inproceedings{jin2024timellm,
  title={Time-{LLM}: time series forecasting by reprogramming large language models},
  author={Jin, Ming and Wang, Shiyu and Ma, Lintao and Chu, Zhixuan and Zhang, James and Shi, Xiaoming and Chen, Pin-Yu and Liang, Yuxuan and Li, Yuan-Fang and Pan, Shirui and others},
  booktitle={International conference on learning representations},
  volume={2024},
  pages={23857--23880},
  year={2024}
}

@article{chan2024medtsllm,
  title={{MedTsLLM}: leveraging {LLM}s for multimodal medical time series analysis},
  author={Chan, Nimeesha and Parker, Felix and Bennett, William and Wu, Tianyi and Jia, Mung Yao and Fackler, James and Ghobadi, Kimia},
  journal={arXiv preprint arXiv:2408.07773},
  year={2024}
}

@article{pillai2025time2lang,
  title={{Time2Lang}: bridging time-series foundation models and large language models for health sensing beyond prompting},
  author={Pillai, Arvind and Spathis, Dimitris and Nepal, Subigya and Collins, Amanda C and Mackin, Daniel M and Heinz, Michael V and Griffin, Tess Z and Jacobson, Nicholas C and Campbell, Andrew},
  journal={arXiv preprint arXiv:2502.07608},
  year={2025}
}

@article{goswami2024moment,
  title={{MOMENT}: a family of open time-series foundation models},
  author={Goswami, Mononito and Szafer, Konrad and Choudhry, Arjun and Cai, Yifu and Li, Shuo and Dubrawski, Artur},
  journal={arXiv preprint arXiv:2402.03885},
  year={2024}
}

@article{wagner2020ptbxl,
  title={{PTB-XL}, a large publicly available electrocardiography dataset},
  author={Wagner, Patrick and Strodthoff, Nils and Bousseljot, Ralf-Dieter and Kreiseler, Dieter and Lunze, Fatima I and Samek, Wojciech and Schaeffter, Tobias},
  journal={Scientific data},
  volume={7},
  number={1},
  pages={154},
  year={2020},
  publisher={Nature Publishing Group UK London}
}

@article{zheng2020chapman,
  title={A 12-lead electrocardiogram database for arrhythmia research covering more than 10,000 patients},
  author={Zheng, Jianwei and Zhang, Jianming and Danioko, Sidy and Yao, Hai and Guo, Hangyuan and Rakovski, Cyril},
  journal={Scientific data},
  volume={7},
  number={1},
  pages={48},
  year={2020},
  publisher={Nature Publishing Group UK London}
}

@article{touvron2023llama2,
  title={{Llama 2}: open foundation and fine-tuned chat models},
  author={Touvron, Hugo and Martin, Louis and Stone, Kevin and Albert, Peter and Almahairi, Amjad and Babaei, Yasmine and Bashlykov, Nikolay and Batra, Soumya and Bhargava, Prajjwal and Bhosale, Shruti and others},
  journal={arXiv preprint arXiv:2307.09288},
  year={2023}
}

@article{zhou2023onefitsall,
  title={One fits all: power general time series analysis by pretrained {LM}},
  author={Zhou, Tian and Niu, Peisong and Wang, Xue and Sun, Liang and Jin, Rong},
  journal={Advances in neural information processing systems},
  volume={36},
  pages={43322--43355},
  year={2023}
}

@article{nie2022patchtst,
  title={A time series is worth 64 words: long-term forecasting with transformers},
  author={Nie, Yuqi and Nguyen, Nam H and Sinthong, Phanwadee and Kalagnanam, Jayant},
  journal={arXiv preprint arXiv:2211.14730},
  year={2022}
}

@article{langer2025opentslm,
  title={{OpenTSLM}: time-series language models for reasoning over multivariate medical text-and time-series data},
  author={Langer, Patrick and Kaar, Thomas and Rosenblattl, Max and Xu, Maxwell A and Chow, Winnie and Maritsch, Martin and Jakob, Robert and Wang, Ning and Liu, Juncheng and Verma, Aradhana and others},
  journal={arXiv preprint arXiv:2510.02410},
  year={2025}
}

@article{lan2026gem,
  title={{GEM}: empowering {MLLM} for grounded {ECG} understanding with time series and images},
  author={Lan, Xiang and Wu, Feng and He, Kai and Zhao, Qinghao and Hong, Shenda and Feng, Mengling},
  journal={Advances in Neural Information Processing Systems},
  volume={38},
  pages={94421--94455},
  year={2026}
}
 }
\end{document}